\documentclass[conference]{IEEEtran}
\IEEEoverridecommandlockouts
\usepackage{cite}
\usepackage{amsmath,amssymb,amsfonts}
\usepackage{algorithmic}
\usepackage{graphicx}
\usepackage{textcomp}
\usepackage{xcolor}
\usepackage{algorithmic}
\usepackage{algorithm}
\usepackage{caption}
\usepackage{caption}
\def\BibTeX{{\rm B\kern-.05em{\sc i\kern-.025em b}\kern-.08em
    T\kern-.1667em\lower.7ex\hbox{E}\kern-.125emX}}

\begin{document}

\title{Co-Design of a Robot Controller Board and Indoor Positioning System for IoT-Enabled Applications \\

}

\author{
\IEEEauthorblockN{Ali Safa$^{1}$, Ali Al-Zawqari$^{2}$}
\IEEEauthorblockA{\textit{$^{1}$College of
Science and Engineering, Hamad Bin Khalifa University, Doha, Qatar},\\
\textit{$^{2}$ELEC Department, Vrije Universiteit Brussel, Brussels, Belgium} \\
asafa@hbku.edu.qa, aalzawqa@vub.be}

}

\maketitle

\begin{abstract}
This paper describes the development of a cost-effective yet precise indoor robot navigation system composed of a custom robot controller board and an indoor positioning system. 
First, the proposed robot controller board has been specially designed for emerging IoT-based robot applications and is capable of driving two 6-Amp motor channels. The controller board also embeds an on-board micro-controller with WIFI connectivity, enabling robot-to-server communications for IoT applications. Then, working together with the robot controller board, the proposed positioning system detects the robot's location using a down-looking webcam and uses the robot's position on the webcam images to estimate the real-world position of the robot in the environment. The positioning system can then send commands via WIFI to the robot in order to steer it to any arbitrary location in the environment. Our experiments show that the proposed system reaches a navigation error smaller or equal to 0.125 meters while being more than two orders of magnitude more cost-effective compared to off-the-shelve motion capture (MOCAP) positioning systems. 
\end{abstract}



\section*{Supplementary Material}
A video demonstration of our proposed system is available at: \texttt{https://tinyurl.com/4vpz5ar7} 

\section{Introduction}

In recent years, the study and design of indoor robot navigation systems has attracted much attention for applications ranging from automated warehouse management to indoor inspection and maintenance \cite{surveyrobots}. In order to conceive the many subsystems needed for indoor robot navigation, research is currently being conducted across different fields such as: \textit{i)} robot mechanics design \cite{scaramuzz}; \textit{ii)} electronics hardware design (e.g., low-power AI processors \cite{memristorprocessor, cmosedge, cnnhardware}, robot controller boards \cite{robotcontroller}); \textit{iii)} motor control algorithms; \textit{iv)} AI-driven perception systems (e.g., people detection and avoidance \cite{peopledetect}, Simultaneous Localisation and Mapping or SLAM \cite{slamdrone}); \textit{v)} Internet of Things (IoT) \cite{iotuav} for robot-to-robot and robot-to-server communication; and \textit{vi)} motion capture (MOCAP) systems for robot positioning \cite{mocap}. 

In this work, our goal is to co-develop a custom robot controller hardware (\textit{ii)} in the list given above) together with an indoor MOCAP positioning system (\textit{vi)} in the list given above), enabling a direct positioning of the robot without resorting to compute-expensive SLAM algorithms. 

\begin{figure}[t]
\centering
    \includegraphics[scale = 0.37]{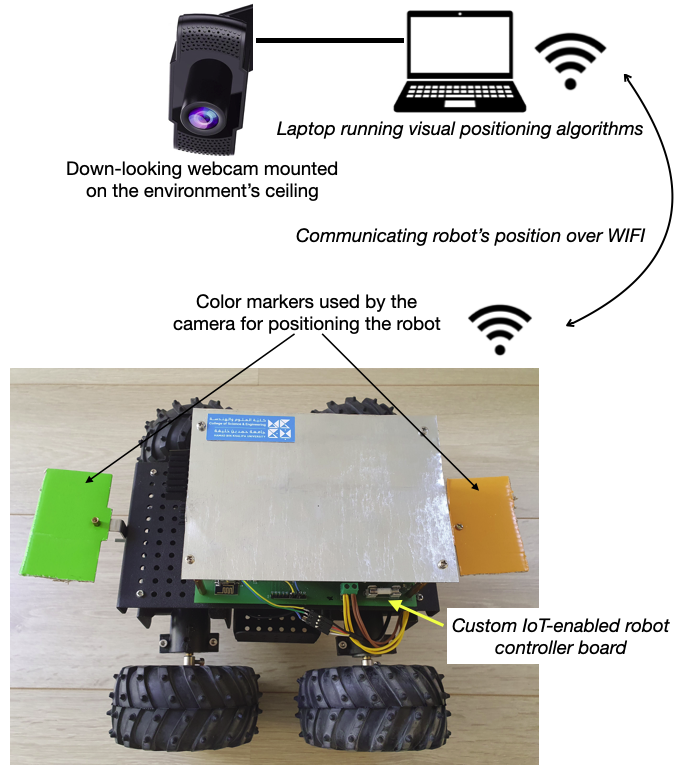}
    \caption{\textit{\textbf{View of the robot controller and indoor positioning system proposed in this work.} A custom robot motor controller board is used to drive the robot's wheels while providing WIFI connectivity for enabling the study of IoT-based applications. Color markers mounted on the robot are used by a down-looking webcam connected to a laptop for visually determining the position of the robot in the environment. The camera-based positioning system transmits the robot's position to the custom controller board via WIFI.}}
    \label{systemview}
\end{figure}

Indeed, the availability of a MOCAP system \cite{mocap, mocap2} is central to most indoor robotics setups as it provides precise positioning to the robot controller system. MOCAP systems typically use an array of infrared cameras mounted on the ceiling above the area to be tracked. Reflective markers are mounted on each robot that needs to be tracked and are detected by the MOCAP camera array. Then, localisation and triangulation algorithms are used to derive the position of each robot using the tracked marker positions \cite{mocap, mocap2}.  

\begin{figure*}[htbp]
\centering
    \includegraphics[scale = 0.34]{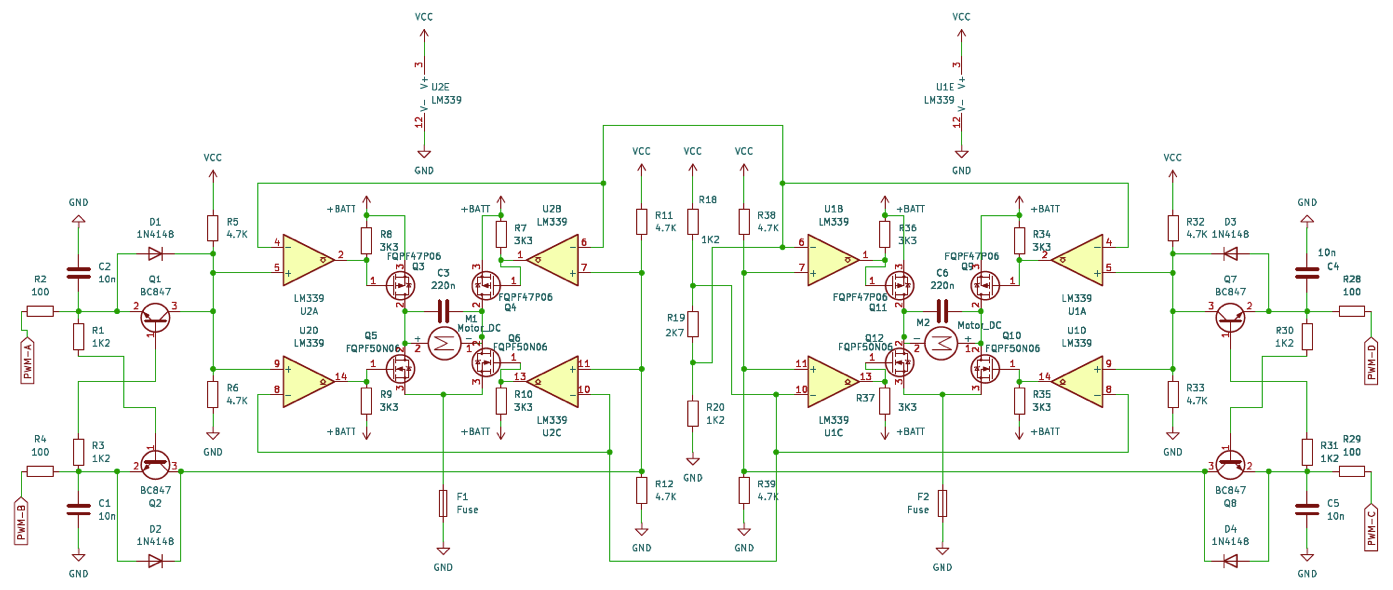}
    \caption{\textit{\textbf{H-bridge circuit design.} The PWM signals for controlling the speed of the right and left motors are fed through the ports \texttt{PWM-A},\texttt{PWM-B}, \texttt{PWM-C} and \texttt{PWM-D}. The motor turns clock-wise when feeding the PWM signal to \texttt{PWM-A} (\texttt{PWM-C}) while keeping \texttt{PWM-B} (\texttt{PWM-D}) to \texttt{GND}, and vice-versa for counter-clock-wise rotation.}}
    \label{hbridgecirc}
\end{figure*}

Even though widely used in \textit{lab settings}, MOCAP systems are known to be highly expensive in terms of purchase costs (with a typical price tag of $\sim$120,000 QAR for covering a $7$-m$\times7$-m area \cite{optitrack}), making them less suited for a cost-effective, \textit{on-the-field} deployment in large areas. 

To help alleviate this issue, this paper describes how to set up a cost-effective yet precise indoor robot navigation system intended to be used for the study of emerging IoT and AI applications (such as e.g., federated learning at the robot edge \cite{federatedlearn}). At the core of the system lies a custom robot controller board with wireless connectivity (through WIFI), providing networking and internet connection capability to the robot platform. Fig. \ref{systemview} provides a ensemble view of the system proposed in this work.

The contributions of this paper are the following:
\begin{enumerate}
    \item We design a robot controller board embarking a WIFI-enabled micro-controller chip providing IoT capability and running a Proportional-Integral-Derivative (PID) motor speed control loop with anti-windup capability.
    \item We show how to set up a cost-effective yet precise indoor positioning system using a down-looking webcam camera connected to a laptop which interacts with the robot controller board through the WIFI connectivity.
    \item We experimentally demonstrate the navigation capability of the proposed system, reaching a low robot positioning error of $\leq 0.125$ meter. 
\end{enumerate}

This paper is organized as follows. The design of our IoT-enabled robot controller is described in Section \ref{robotcontrol}. Our proposed indoor robot positioning system is detailed in Section \ref{indoorpos}. Demonstrations about the robot navigation capability of the proposed setup are provided in Section \ref{experiments}. Finally, conclusions are provided in Section \ref{conclusion}.

\section{IoT-enabled robot controller design}
\label{robotcontrol}

\begin{figure}[htbp]
\centering
    \includegraphics[scale = 0.35]{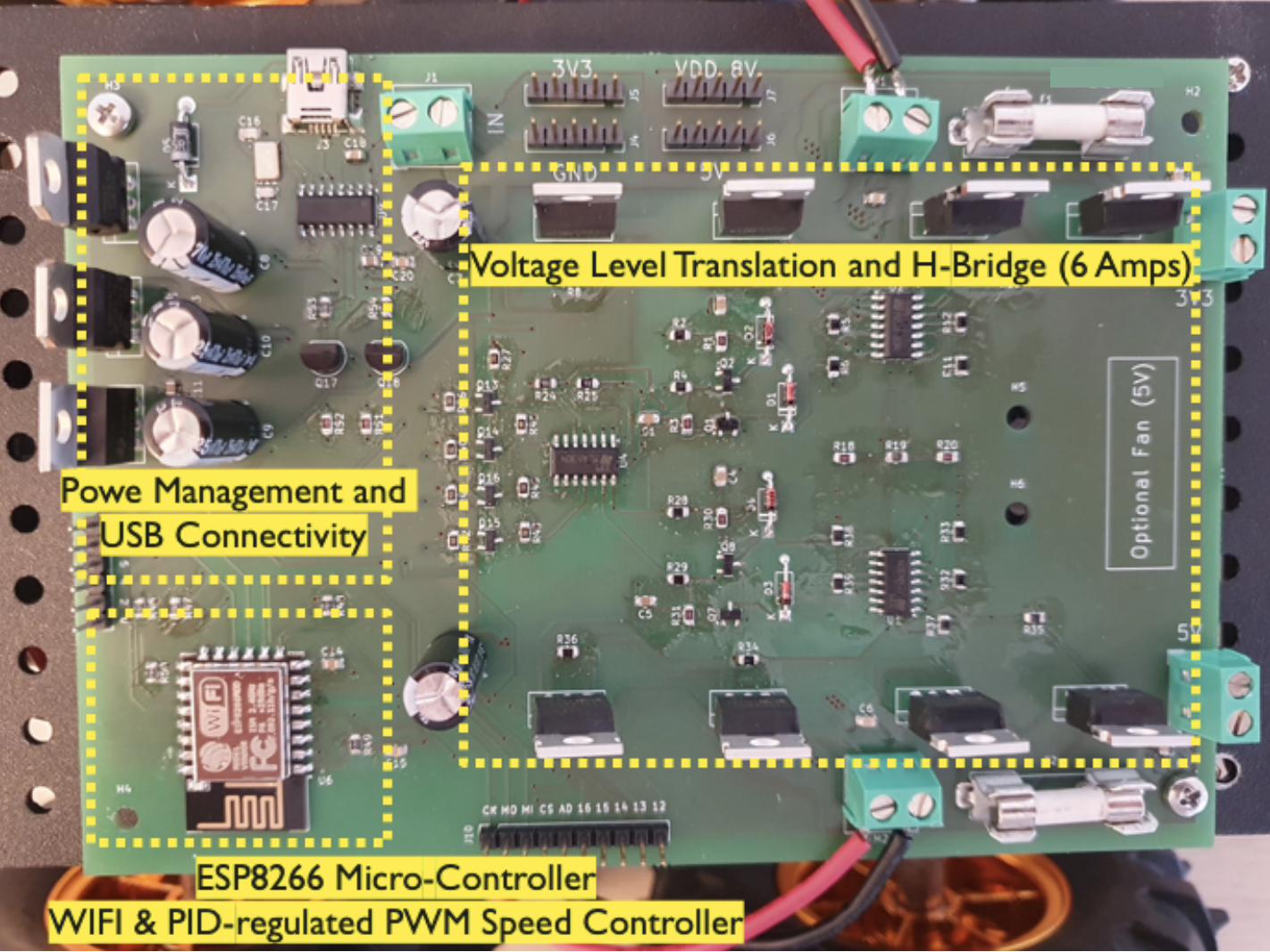}
    \caption{\textit{\textbf{Custom robot controller board.} The board embarks a H-bridge circuit capable of driving motors with up to 6 Amps of current. The H-bridge is controlled via PWM by the on-board ESP8266 micro-controller chip which also provides WIFI connectivity to the board.}}
    \label{motorcontroller}
\end{figure}

In order to control the motor speed of the rover bot used in this work (see Fig. \ref{systemview}) while providing wireless connectivity to the robot system, the custom controller board in Fig. \ref{motorcontroller} has been designed, embarking a 6-Amp H-bridge circuit \cite{hbridge}, a WIFI-enabled ESP8266 micro-controller chip, various power management circuits ($3.3$-V, $5$-V and $8$-V regulation) and a USB to serial conversion interface (using a CH340 chip).

The ESP8266 can be programmed via the USB interface and is used to control the speed of the robot's motors via Pulse-Width Modulation (PWM) applied to the gates of the MOSFETs used within the H-bridge circuit (see Fig. \ref{hbridgecirc}). 

\begin{figure}[htbp]
\centering
    \includegraphics[scale = 0.18]{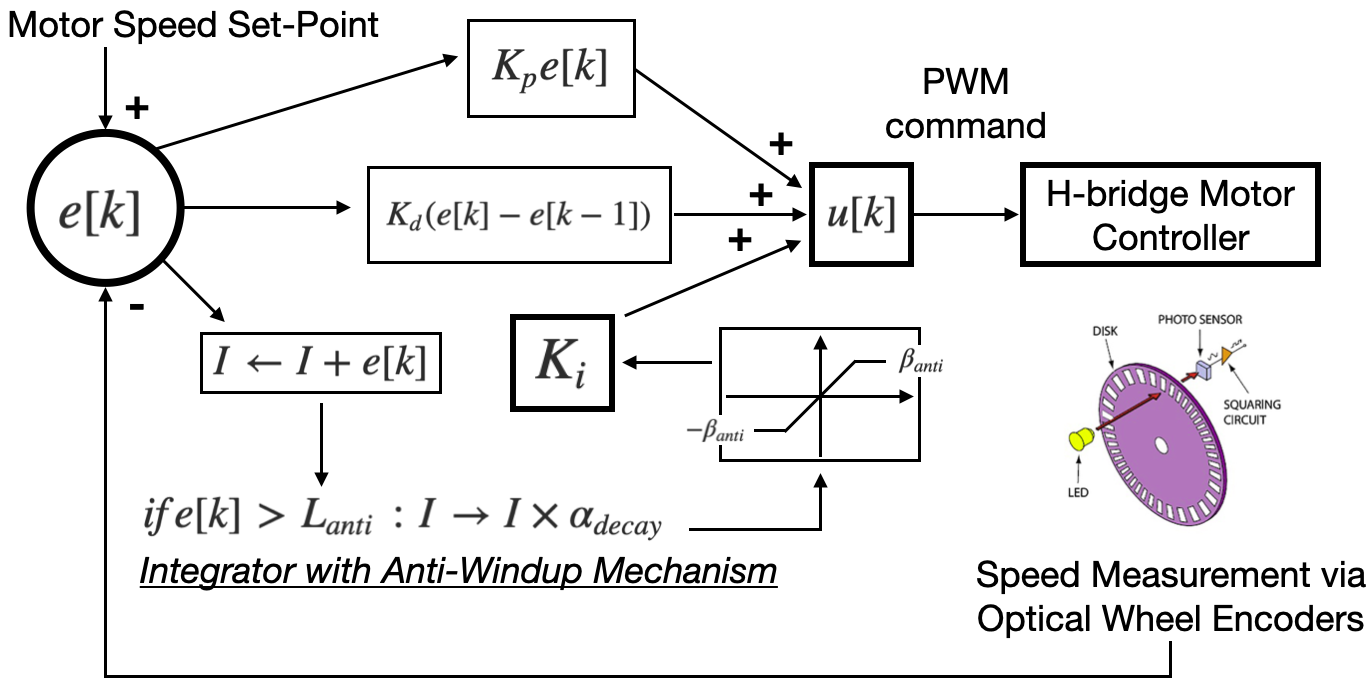}
    \caption{\textit{\textbf{Anti-windup PID motor speed controller.} The error signal $e[k]$ at time-step $k$ is obtained by subtracting the measured rotation speed from the desired set-point speed. The error signal $e[k]$ is then fed to a proportional, a derivative and an integral pipeline with coefficients $K_p, K_d, K_i$. The anti-windup mechanism applies a decay $I\xleftarrow{} I\times \alpha_{decay}$ to the integrator if the error is larger than a certain threshold $L_{anti}$. The integrator is further clamped between $(-\beta_{anti}, \beta_{anti})$ to prevent large bursts in the PWM commands.}}
    \label{antiwindup}
\end{figure}

The PWM motor speed control loop running in the ESP8266 is built around a PID controller \cite{PIDantiwindup} and uses the motor speed measurement from the optical wheel encoders as input to the PID controller \cite{cooperativerobot}. The desired speed set-point is compared to the measured rotation speed in order to derive the error signal used by the PID controller to adjust its PWM command to the motors. The PID controller is also equipped with an anti-windup mechanism \cite{PIDantiwindup} preventing the integral error to grow indefinitely in case the wheels are being stuck due to the floor conditions. Fig. \ref{antiwindup} shows the anti-windup PID control loop running in the ESP8266. In addition, Table \ref{paramspid} reports the PID and anti-windup control parameters used in this work (tuned empirically).

\begin{table}[htbp] 
\centering
\begin{tabular}{|c|c|c|c|c|c|c|}
\hline
$K_p$ & $K_d$ & $K_i$ & $L_{anti}$ & $\alpha_{decay}$ & $\beta_{anti}$ \\
\hline
$1.2$ & $0.05$ & $3$ & $10$ & $0.4$ & $100$ \\
\hline
\end{tabular}
\caption{\textit{\textbf{PID and anti-windup control parameters.} The control parameters are described in Fig. \ref{antiwindup}. The PID operates at a rate of $10$-Hz.}}
\label{paramspid}
\end{table}

\section{Indoor Positioning System}
\label{indoorpos}
\begin{figure}[t]
\centering
    \includegraphics[scale = 0.3]{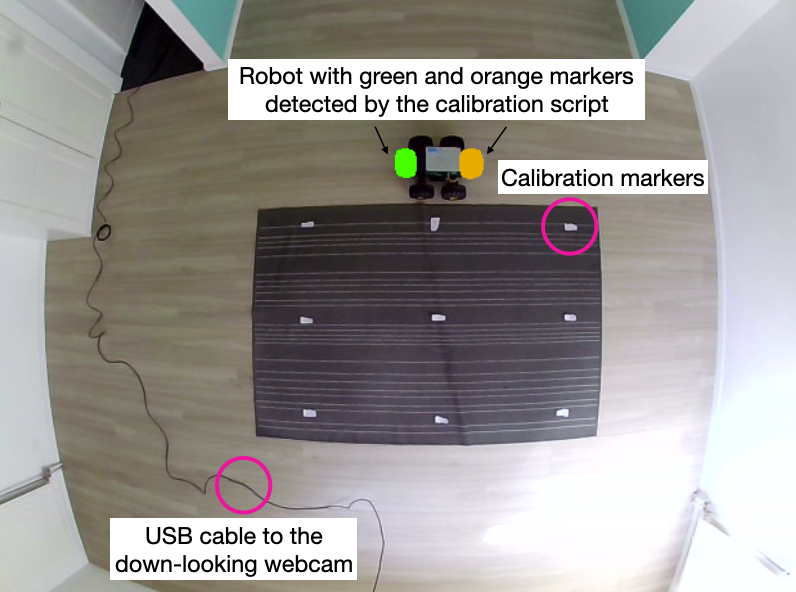}
    \caption{\textit{\textbf{Calibration setup.} The white markers placed on the floor indicate the location of the calibration points (with known real-world coordinate $(X_c^i,Y_c^i)$). During calibration, the robot is placed on top of a marker and its location on the camera image plane $(x_c^i,y_c^i)$ is determined by detecting the green and orange markers mounted on the robot (see Fig. \ref{systemview}).}}
    \label{calibration_setup}
\end{figure}

\begin{figure*}[htbp]
\centering
    \includegraphics[scale = 0.34]{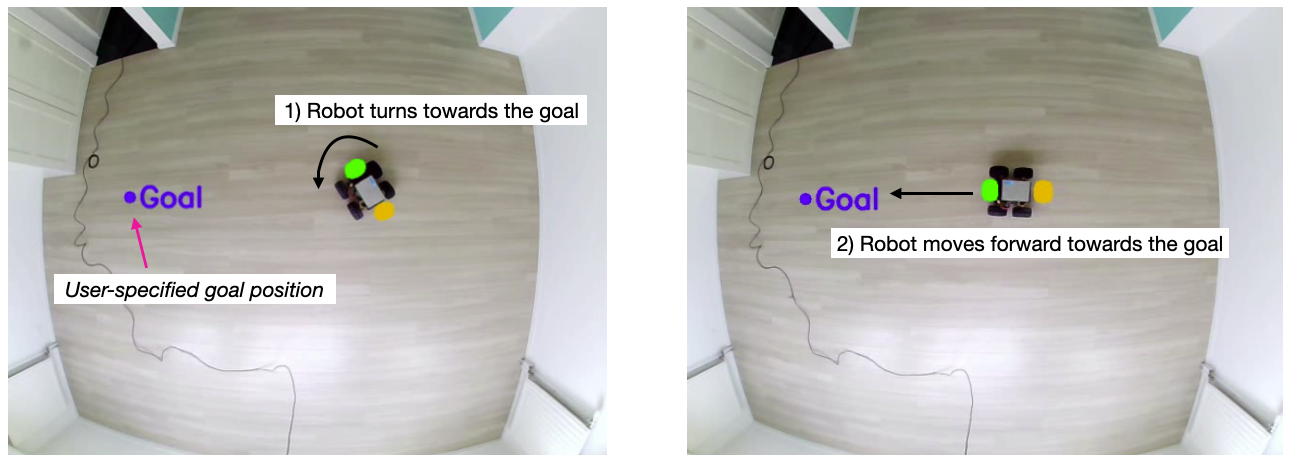}
    \caption{\textit{\textbf{Experiment 1: Steering the robot towards a user-specified goal.} First, the user specifies a goal position to be reached by clicking on the desired location on the image plane. Upon receiving the user-specified location, the python script running the visual positioning algorithm described in Section \ref{indoorpos} computes the real-world goal and robot coordinates, and sends commands to the robot via WIFI (see Fig. \ref{systemview}) in order to steer it towards the user-specified goal location. The steering is done by 1) turning the robot towards the goal position and 2) moving forward until the goal is reached.  }}
    \label{steeringpic}
\end{figure*}

We set up an inexpensive yet effective indoor positioning system by mounting a down-looking webcam to the ceiling of the indoor environment in which our robot is set to navigate. We choose a wide-angle camera (HBV-1716WA 2MP camera) with a field of view of $140^\circ$ in order to maximise the floor coverage area. By mounting the green and orange color markers on the robot chassis (see Fig. \ref{systemview}), it is possible to determine both the robot location $(x_c,y_c)$ and azimuth orientation angle $\theta_c$ on the image plane of the webcam as follows. 

The webcam is connected to a laptop running a real-time python script which localizes all the pixel coordinates $\{(x^g_i, y^g_i), \forall i = 1,n_g\}$ corresponding to the distinct green marker, and all the pixel coordinates $\{(x^o_j, y^o_j), \forall j=1,n_o\}$ corresponding to the distinct orange marker. Using these pixel coordinates, it is possible to find the centroid coordinates $(x^g_c, y^g_c); (x^o_c, y^o_c)$ of the green and orange markers via (\ref{centroid1}) and (\ref{centroid2}) respectively.
\begin{equation}
    (x^g_c, y^g_c) = \frac{1}{n_g} \sum_{i=1}^{n_g} (x^g_i, y^g_i)
    \label{centroid1}
\end{equation}

\begin{equation}
    (x^o_c, y^o_c) = \frac{1}{n_o} \sum_{j=1}^{n_o} (x^o_j, y^o_j)
    \label{centroid2}
\end{equation}

Then, the location of the robot on the image plane $(x_c, y_c)$ can be derived as the average location between the green and orange centroids:
\begin{equation}
    (x_c, y_c) = \frac{(x^g_c, y^g_c) + (x^o_c, y^o_c)}{2} 
    \label{centroid3}
\end{equation}

In addition, the azimuth orientation angle of the robot $\theta_c$ can be derived as:
\begin{equation}
    \theta_c = \arctan \frac{x^o_c - x^g_c}{y^o_c - y^g_c} 
    \label{azimuth}
\end{equation}

Now, in order to derive the robot location $(X_c,Y_c)$ in terms of real-world coordinates (vs. coordinates on the image plane), we use the equations of the pinhole camera \cite{pinholecam} to relate the real-world and pixel coordinates together:
\begin{equation}
    (x_c,y_c) = \frac{f}{Z} (X_c - C_x,Y_c - C_y)
    \label{pinhole}
\end{equation}
where $(C_x,C_y)$ is the origin of the real-world coordinate system, $f$ is the camera's focal length and $Z$ is the \textit{constant} height distance from the robot to the down-looking camera mounted to the ceiling \cite{pinholeca2}. The four unknowns $(C_x,C_y)$, $f$ and $Z$ can be determined through \textit{calibration}, by 1) placing the robot in \textit{at least} four different \textit{known} positions $\{(X_c^i,Y_c^i), i= 1,2,3,4\}$ and acquiring their corresponding image plane positions $\{(x_c^i,y_c^i), i= 1,2,3,4\}$; and 2) solving a system of four equations to determine the four unknowns $(C_x,C_y)$, $f$ and $Z$.

Fig. \ref{calibration_setup} shows the calibration setup used in this work. During the calibration process, the robot is placed on the different calibration points with real-world coordinate $(X_c^i,Y_c^i)$ distributed on the floor (see the white markers in Fig. \ref{calibration_setup}). Then, the centroid locations of the green and orange markers mounted on the robot are determined in order to compute the location of the calibration points $(x_c^i,y_c^i)$ \textit{on the image plane}. Finally, the acquired points can be used to compute the camera parameters $(C_x,C_y)$, $f$ and $Z$ following the calibration process described in this Section.



Once $(C_x,C_y)$, $f$ and $Z$ have been determined, it is possible to use (\ref{pinhole}) to derive the real-world robot position $(X_c,Y_c)$. Note that the azimuth orientation angle $\theta_c$ stays identical when passing from the image plane to the real-world coordinate system since the webcam is mounted horizontally with its image plane parallel to the ground plane.

Finally, the python script running the positioning algorithm can \textit{stream} via WIFI the real-world position of the robot to the robot's custom controller board described in Section \ref{robotcontrol}. Doing so, a complete IoT-enabled indoor robot navigation setup can be formed and used to study future robotics applications equipped with wireless networking capability (see Fig. \ref{systemview}).



\section{Robot navigation experiments}
\label{experiments}

\subsection{Experiment 1: Steering the robot towards a goal}
In order to demonstrate the navigation capability of our proposed system in Fig. \ref{systemview}, we set up a first experiment where the goal is to steer the robot from an arbitrary start position to an arbitrary end position. 

To select the arbitrary end goal $(X_{goal}, Y_{goal})$, an external user specifies the goal location to be reached by clicking on the desired image plane location from the real-time webcam video feed. Then, the python script running the positioning system (described in Section \ref{indoorpos}) transforms this user-specified image plane location into a real-world coordinate $(X_{goal}, Y_{goal})$.

In order to steer the robot from its starting position $(X_c,Y_c)$ to the goal position $(X_{goal}, Y_{goal})$, we adopt a 2-step steering scheme which works as follows:
\begin{enumerate}
    \item First, the azimuth orientation angle $\theta_c$ of the robot gets aligned with the goal location, by aligning $\theta_c$ with the line joining $(X_c,Y_c)$ and $(X_{goal}, Y_{goal})$. The python script running the positioning system sends a \texttt{turn} command via WIFI to the robot which makes the robot right and left wheels turn in \textit{opposite} directions at an angular speed of $10$ rad/s. Once $\theta_c$ gets aligned with goal, the python script sends a \texttt{stop} command which halts the robot.
    \item Then, the robot is driven forward until the goal is reached. This is done by sending a \texttt{forward} command via WIFI to the robot which makes the robot wheels turn in the \textit{same} direction at an angular speed of $10$ rad/s. Once $(X_c,Y_c)$ reaches the desired goal $(X_{goal}, Y_{goal})$ within an acceptance perimeter of $\sqrt{(X_c - X_{goal})^2 + (Y_c - Y_{goal})^2} < 0.1$ meters, a \texttt{stop} command is sent via WIFI which halts the robot.
\end{enumerate}

Fig. \ref{steeringpic} illustrates the goal selection and steering scheme described above. Furthermore, Table \ref{prcisiontable} reports the positioning precision of our proposed system, together with its standard deviation. The precision is computed by averaging the errors obtained between the goal position $(X_{goal}, Y_{goal})$ and the final position reached by the robot over 50 trials (with randomly-selected goal positions).
\begin{table}[htbp] 
\centering
\begin{tabular}{|c|c|}
\hline
\textbf{Average positioning error [m]} & \textbf{Standard deviation} [m] \\
\hline
0.125 & 0.0439  \\
\hline
\end{tabular}
\caption{\textit{\textbf{Positioning error and standard deviation.} Computed over 50 randomly-selected goal positions. }}
\label{prcisiontable}
\end{table}

\subsection{Experiment 2: Trajectory tracking}

In our second series of experiments, we consider the more challenging case of \textit{trajectory tracking}, where the robot must follow a precise user-defined trajectory with as little deviation as possible.

\begin{figure*}[htbp]
\centering
    \includegraphics[scale = 0.4]{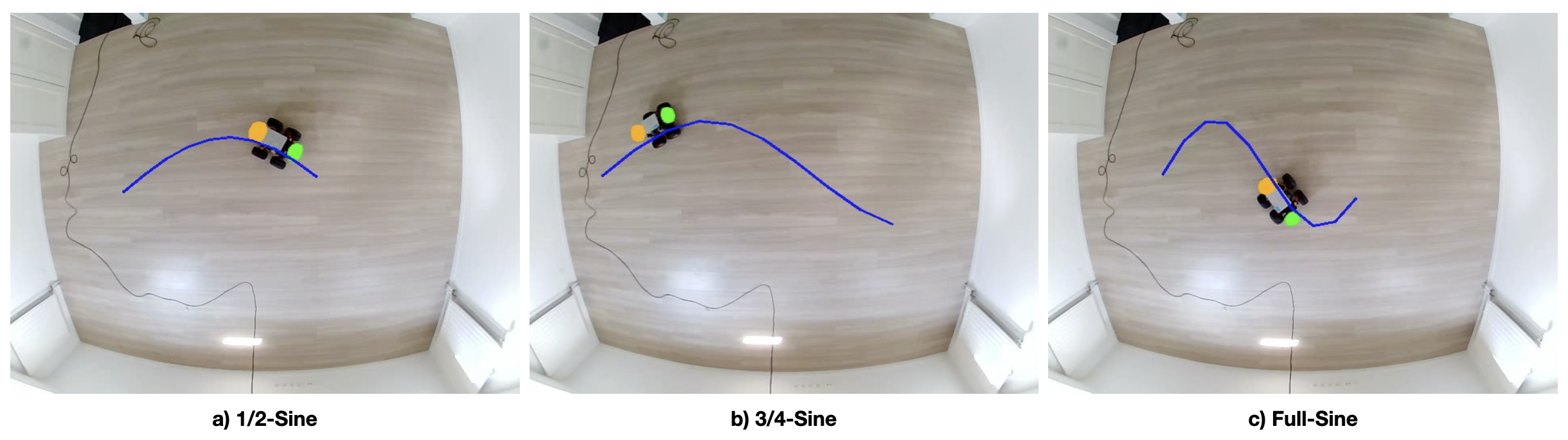}
    \caption{\textit{\textbf{Experiment 2: Trajectory tracking.} a) the robot follows a trajectory spanning half a sine wave. b) the robot follows a trajectory spanning three-quarter of a sine wave. c) the robot follows a full sine wave.}}
    \label{trajectorytracking}
\end{figure*}

\begin{figure}[htbp]
\centering
    \includegraphics[scale = 0.3]{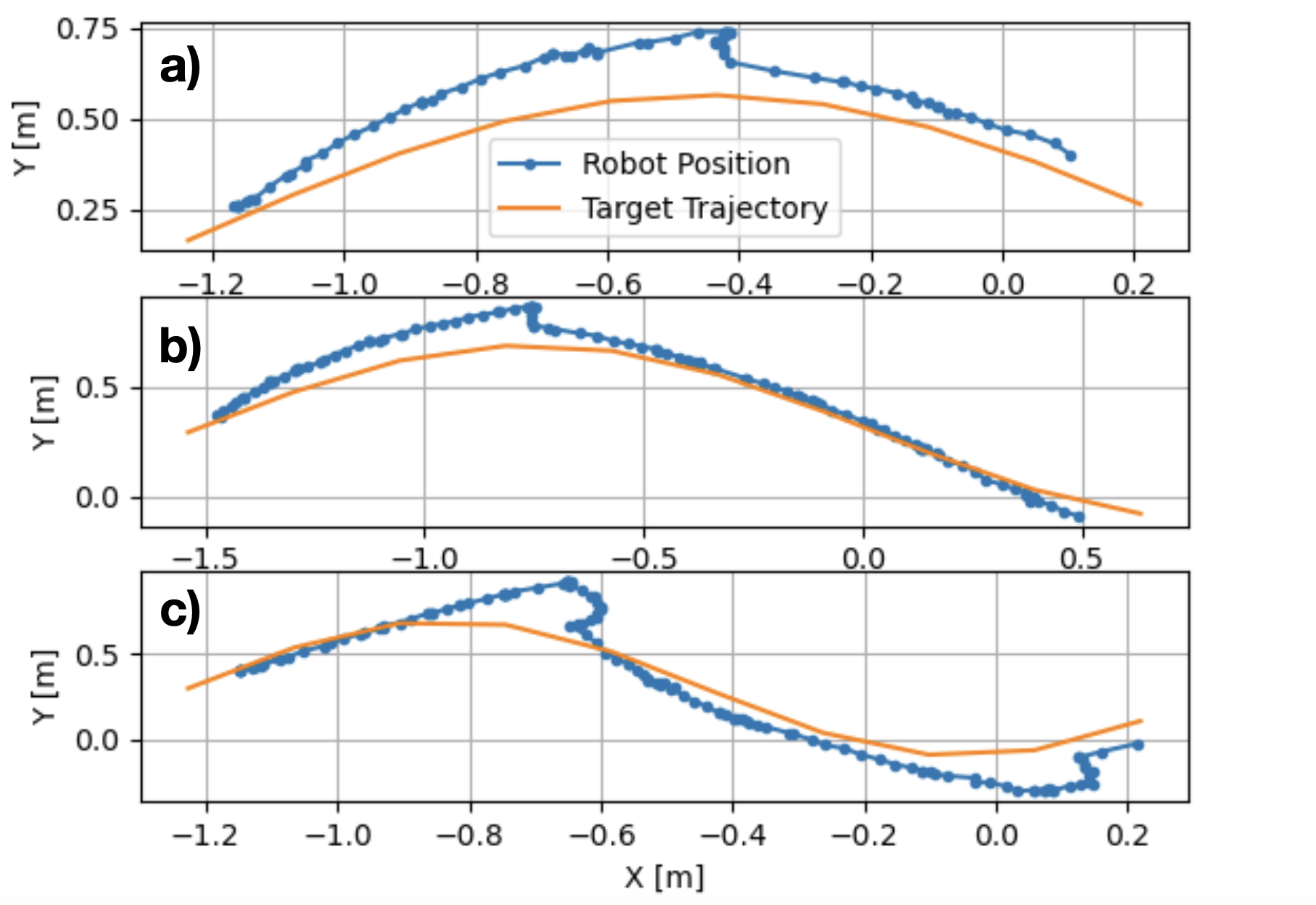}
    \caption{\textit{\textbf{Real-world robot position vs. the target trajectory to be tracked.} }}
    \label{trajerrors}
\end{figure}

We experiment with three different trajectories shown in Fig. \ref{trajectorytracking}. This trajectories are all generated by spanning either half, either three-quarter or either a full period of a sine wave. Furthermore, the \textit{orange curves} in Fig. \ref{trajerrors} show the target trajectory curves in the real-world coordinate system.

To perform tracking, we consider a control scheme built around a Proportional regulator (tuned empirically) which steers the orientation of the robot back towards the track in function of the error distance $\Delta d$ between the robot's green marker and its closest point on the track. In addition, when this error $\Delta d$ crosses a threshold $\theta_d$, the robot executes a rotation with no forward translation for that time step. We found this additional control rule helpful for cases where the track curvature is important (e.g., near the maxima of the Full-Sine trajectory in Fig. \ref{trajectorytracking} c). Algorithm \ref{trackersch} details the tracker control loop, which sends velocity commands to the anti-windup PID regulator of our robot controller board (see. Fig. \ref{antiwindup}).
 \begin{algorithm}
 \caption{Trajectory tracker control loop}
 \label{trackersch}
 \begin{algorithmic}[1]
 \renewcommand{\algorithmicrequire}{\textbf{Input:}}
 \renewcommand{\algorithmicensure}{\textbf{Output:}}
 \REQUIRE $\bar{s}=\{(X_k^*, Y_k^*) \} \forall k$: Target trajectory to be tracked.
  \REQUIRE $\theta_d = 0.15$: deviation threshold in m.
  \REQUIRE $v_{m} = 0.5$: \textit{baseline} motor speed in rad/s.
  \REQUIRE $K_{p}^{t} = 70$: Proportional coefficient.

  \WHILE{goal not reached}
  \STATE // $(X, Y)$ is the current robot position.
  \STATE $\Delta d \xleftarrow{} \min_k ||\bar{s}_k - (X, Y)||_2$ 
  \IF{$(X, Y)$ is on the \textbf{right} side of the track}
    \STATE $\delta = 1$
  \ELSIF{$(X, Y)$ is on the \textbf{left} side of the track}
    \STATE $\delta = -1$
  \ENDIF

  \STATE $u_{\text{right}} = v_{m} - \delta \times K_{p}^{t} \times \Delta d$ // Proportional regulation
    \STATE $u_{\text{left}} = v_{m} + \delta \times K_{p}^{t} \times \Delta d$

    \IF{$\Delta d > \theta_d$} 
    \STATE $u_{\text{right}} = -\delta \times v_{m}$ //Over-write the right and left speeds
    \STATE $u_{\text{left}} = \delta \times v_{m}$ // to make a pure rotation
    \ENDIF

  \STATE Send right and left target motor speeds $u_{\text{right}}, u_{\text{left}}$ to the robot via WIFI.
  \ENDWHILE
  
 \end{algorithmic} 
 \end{algorithm}

Finally, Table \ref{prcisiontable2} reports the average trajectory deviation errors for the three sine wave tracking experiments in Fig. \ref{trajerrors}.
\begin{table}[htbp] 
\centering
\begin{tabular}{|c|c|c|}
\hline
\textbf{1/2-Sine} & \textbf{3/4-Sine} & \textbf{Full-Sine} \\
\hline
0.1 m & 0.068 m & 0.106 m\\
\hline
\end{tabular}
\caption{\textit{\textbf{Averaging trajectory deviation error during the tracking experiments.} }}
\label{prcisiontable2}
\end{table}

\subsection{Discussion}
We observe that the positioning errors ($\leq 0.125$ m) reported in Tables \ref{prcisiontable} and \ref{prcisiontable2} are \textit{smaller} compared to the physical robot dimensions ($0.3$ m $\times 0.2$ m), making the proposed systems usable in practice. This clearly demonstrates the usefulness of our proposed indoor robot navigation system, while costing more than \textit{two orders of magnitude less} compared to off-the-shelve positioning and MOCAP systems \cite{optitrack} ($\sim 1600$ QAR for our proposed system vs. $\sim 120,000$ QAR for off-the-shelve MOCAPs). Even though the robot always reached the user-specified goal and tracked the desired trajectories with acceptable precision during the experiments, it must be noted that the precision of the proposed navigation system could be enhanced even more by using more advanced robot steering schemes such as e.g., Linear-Quadratic (LQ) controllers \cite{diffwheeledrobot}. Finally, a video showing how our proposed system operates is provided as supplementary material at \texttt{https://tinyurl.com/4vpz5ar7}.

\section{Conclusion}
\label{conclusion}

This paper has described the development of a cost-effective yet precise indoor robot navigation system composed of a custom robot controller board and a camera-based indoor positioning system, communicating together via WIFI. 
The indoor positioning setup and the controller board have been used to demonstrate the robot navigation capability of the proposed system. It has been shown that our system achieves a positioning error $\leq 0.125$ meter while being \textit{two orders of magnitude} less expensive compared to off-the-shelve MOCAP systems. We hope that the developments proposed in this paper will be useful to other researchers during the conception of their own indoor robot navigation setups.

\end{document}